\documentclass{article}
\usepackage{ijcai26}

\usepackage{times}
\usepackage{soul}
\usepackage{url}
\usepackage[hidelinks]{hyperref}
\usepackage[utf8]{inputenc}
\usepackage[small]{caption}
\usepackage{graphicx}
\usepackage{amsmath}
\usepackage{amsthm}
\usepackage{booktabs}
\usepackage{algorithm}
\usepackage{algorithmic}
\usepackage[switch]{lineno}
\usepackage{xspace}
\usepackage{amsfonts}
\usepackage{bm}
\usepackage{multirow}
\usepackage[mode=tex,obeyclassoptions=true]{standalone}
\usepackage{graphicx}
\usepackage{tabularx}
\usepackage{subcaption}
\usepackage{makecell}
\usepackage{tikz}

\usetikzlibrary{
  arrows.meta,
  positioning,
  fit,
  calc,
  shapes.geometric,
  backgrounds,
  decorations,
  decorations.markings
}

\newlength{\omniLineWidth}
\newlength{\omniThickLineWidth}
\newlength{\omniArrowLineWidth}
\newlength{\omniItemPad}
\newlength{\omniItemPadTop}
\newlength{\omniWrapperPad}
\newlength{\omniWrapperPadTop}
\newlength{\omniWrapperTextPad}
\newlength{\omniScopePad}
\newlength{\omniArrowPad}
\newlength{\omniArrowSpacing}
\newlength{\omniItemCorner}
\newlength{\omniWrapperCorner}
\newlength{\omniOuterCorner}
\newlength{\omniItemHeight}
\newlength{\omniItemDoubleHeight}
\newlength{\omniItemWidth}
\newlength{\omniTwoColumnWidth}
\newlength{\omniTwoRowHeight}
\newlength{\omniThreeRowHeight}
\newlength{\omniSingleColumnWidth}
\newlength{\msLineWidth}
\newlength{\msThickLineWidth}
\newlength{\msItemCorner}
\newcommand{\MarkX}{{\textcolor{red!70!black}{\bfseries X}}}
\newcommand{\MarkO}{{\textcolor{blue!70!black}{\bfseries O}}}
\newcommand{\leftMark}[2]{\csname leftMark#1#2\endcsname}
\expandafter\def\csname leftMark00\endcsname{}\expandafter\def\csname leftMark01\endcsname{}\expandafter\def\csname leftMark02\endcsname{}
\expandafter\def\csname leftMark10\endcsname{}\expandafter\def\csname leftMark11\endcsname{}\expandafter\def\csname leftMark12\endcsname{}
\expandafter\def\csname leftMark20\endcsname{}\expandafter\def\csname leftMark21\endcsname{}\expandafter\def\csname leftMark22\endcsname{}

\def\rightGridBG{white}
\newcommand{\rightCellBG}[2]{\csname rightCellBG#1#2\endcsname}
\expandafter\def\csname rightCellBG00\endcsname{\rightGridBG}
\expandafter\def\csname rightCellBG01\endcsname{\rightGridBG}
\expandafter\def\csname rightCellBG02\endcsname{\rightGridBG}
\expandafter\def\csname rightCellBG10\endcsname{\rightGridBG}
\expandafter\def\csname rightCellBG11\endcsname{\rightGridBG}
\expandafter\def\csname rightCellBG12\endcsname{\rightGridBG}
\expandafter\def\csname rightCellBG20\endcsname{\rightGridBG}
\expandafter\def\csname rightCellBG21\endcsname{\rightGridBG}
\expandafter\def\csname rightCellBG22\endcsname{\rightGridBG}
\newcommand{\SetRightCellBG}[3]{\expandafter\def\csname rightCellBG#1#2\endcsname{#3}}

\newcommand{\totalBG}[1]{\csname totalBG@#1\endcsname}
\expandafter\def\csname totalBG@rowSum0\endcsname{white}
\expandafter\def\csname totalBG@rowSum1\endcsname{white}
\expandafter\def\csname totalBG@rowSum2\endcsname{white}
\expandafter\def\csname totalBG@colSum0\endcsname{white}
\expandafter\def\csname totalBG@colSum1\endcsname{white}
\expandafter\def\csname totalBG@colSum2\endcsname{white}
\expandafter\def\csname totalBG@diagSum\endcsname{white}
\expandafter\def\csname totalBG@leftBottomSum\endcsname{white}
\newcommand{\SetTotalBG}[2]{\expandafter\def\csname totalBG@#1\endcsname{#2}}

\newcommand{\SetLeftMark}[3]{\expandafter\def\csname leftMark#1#2\endcsname{#3}}

\newlength{\memLineWidth}
\newlength{\memThickLineWidth}
\newlength{\memArrowLineWidth}
\newlength{\memInnerPad}
\newlength{\memOuterPad}
\newlength{\memBiggerOuterPad}
\newlength{\memInnerCorner}
\newlength{\memOuterCorner}
\newlength{\memItemHeight}
\newlength{\memItemWidth}
\newlength{\memVerticalSpacing}
\newlength{\memHorizontalSpacing}
\newlength{\memEngineWidth}
\input{./shared}

\newcommand{\urloranon}[1]{\url{#1}}

\newcommand{\refsection}[1]{Section~\ref{#1}}
\newcommand{\reftable}[1]{Table~\ref{#1}}
\newcommand{\reffigure}[1]{Figure~\ref{#1}}

\newcommand{\refsupp}[1]{Appendix~\ref{#1}}
\newcommand{\refthreesupps}[3]{Supplementaries~\ref{#1}, \ref{#2}, and \ref{#3}}

\newcommand{\rci}[3]{%
  \makecell[c]{#1\\[-2pt]{\scriptsize[#2, #3]}}%
}

\newcommand{\mci}[2]{%
  \makecell[c]{#1\\[-2pt]{\scriptsize$\pm$#2}}%
}

\title{Towards Improving Sequential Decision-Making in LLM Agents via Experience Memory}

\author{
Jakub Rada
\and
Viliam Lisý
\affiliations
AI Center, Department of Computer Science, FEE, Czech Technical University in Prague\\
\emails
radajak5@fel.cvut.cz,
lisyvili@fel.cvut.cz
}

\begin{document}

\maketitle

\begin{abstract}
Large language models have improved substantially on single-shot reasoning tasks, but their performance in sequential decision-making is less well understood.
We study this on fully-observable two-player zero-sum games, which provide ground-truth evaluation: outcomes are determined by the rules, and optimality of individual moves can be computed or approximated, without relying on a judge model.
Across model tiers, LLMs play suboptimally in simple games such as \ttt or \cfour, and lose to MCTS opponents.
Obfuscations that preserve the game tree but rewrite its surface form leave performance largely unchanged, indicating the gap is not fully explained by recall of memorized strategies.
Motivated by this performance gap, we introduce an agentic framework enhanced with an experience memory designed for the sequential setting and addressing common challenges of sequential decision-making such as credit assignment.
We show that post-game reflection and rule extraction yield measurable improvements on \ttt without modifying the model weights.
\end{abstract}

\begin{figure*}[!t]
  \centering
  \resizebox{0.95\linewidth}{!}{%
    \input{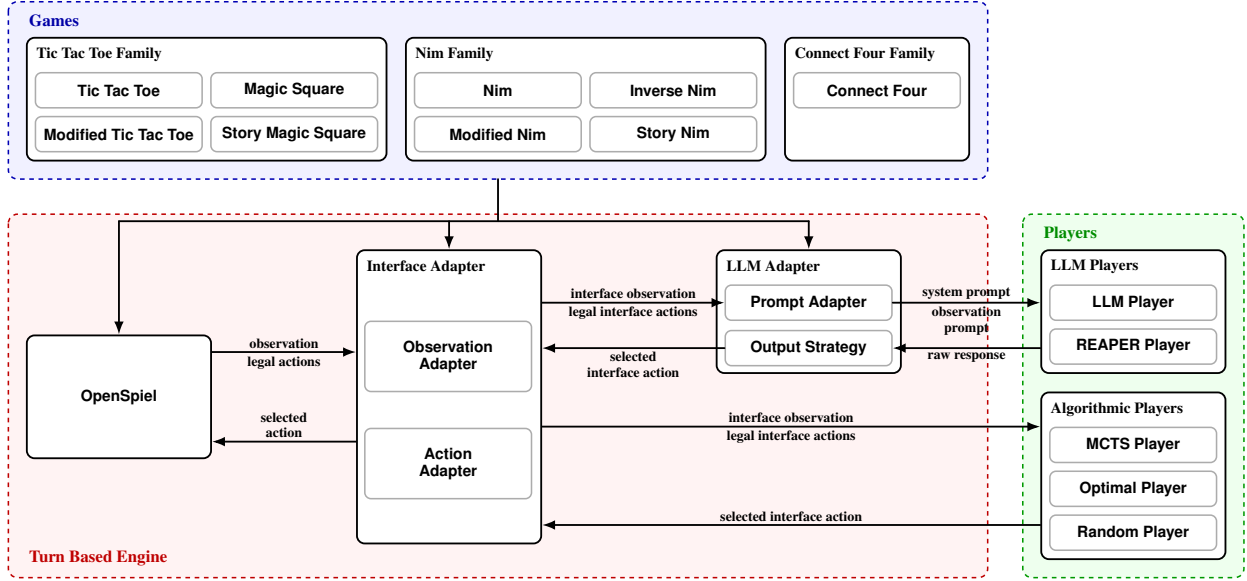}
  }
  \caption{A simplified schema of \omniplay highlighting the modularity and extensibility of the framework to add new games and players.}
  \label{fig:evals:omniplay}
\end{figure*}

\section{Introduction}\label{sec:intro}
Large language models (LLMs) now solve impressive reasoning problems, from competition mathematics to multi-file coding tasks.
Yet a growing body of evidence, most visibly the recent study on the limits of "thinking" models~\cite{illusionofthinking}, shows that this competence does not transfer cleanly to simple sequential decision-making puzzles.
Even on tasks humans master with little practice, frontier models falter once the environment is no longer stationary or passive: it evolves between steps, an adversary pushes back, and rewards are delayed.
Closing this gap matters for deploying LLMs as reliable autonomous agents, yet progress is hard to measure, as benchmarks often lean on LLM judges or proxy metrics that conflate reasoning quality with surface plausibility~\cite{justiceorprejudice}.

We study this gap in a setting that admits ground-truth evaluation: deterministic perfect-information two-player zero-sum games.
Outcomes are determined by the rules, the optimality of each individual move can be computed exactly by classical solvers for games of modest size, and the games we consider, \ttt, \nim, and \cfour, are simple enough that a casual human player can master them.
Building on GTBench~\cite{gtbench}, we evaluate \sota models from OpenAI, Google, and the open-source community across three capability tiers.
The picture is consistent and discouraging.
\efficient and \intermediate models miss four to twelve optimal moves per hundred on \ttt and up to five on \nim, and the \frontier tier, \gpt, \gemini, and \glm, lose to a moderate MCTS opponent on \cfour in essentially every game, typically collapsing within ten moves of a forty-move game.
To probe how robust this behavior is, we evaluate the same models under a suite of game-rule obfuscations that preserve the underlying game tree but rewrite its surface form.
Performance is largely invariant across formulations, and reasoning traces frequently identify the underlying game by name even with disguised rules.
The obfuscations therefore do not cleanly separate reasoning from recall, models can still draw on memorized strategies once they recognize the game, but they do show that the failures are stable properties of how each model engages with these positions, not artifacts of a particular representation.

A natural response would be to fine-tune the models on game play, but updating frontier-model weights is costly, often inaccessible, and risks regressing the capabilities the agent depends on for parsing rules and planning over them.
We therefore ask whether the gap can be closed at the agent level, by supplying the missing decision-time signal externally and leaving the model untouched.
Prior work on case-based reasoning and self-reflection, Memento~\cite{memento}, ExpeL~\cite{expel}, and Reflexion~\cite{reflexion}, has shown that experience memory and generating reflective feedback can yield consistent gains on question-answering and tool-use tasks without weight updates.
These methods, however, were designed for planning rather than adversarial environments; they do not address the challenge of assigning a delayed reward across a sequence of turn-based, adversarial state evolutions.
A direct port to game play is essentially non-functional in our experiments.

We introduce \recap (\recapfull),\footnote{It learns only from its own experience, i.e. it \textit{reaps what it sows}.} a reflective case-based reasoning framework that adapts experience memory to sequential play.
\recap keeps the LLM as the primary reasoner and targets the structural challenges of the setting with a self-reflecting mechanism that performs per-move credit assignment after each completed trajectory. It emits fine-grained annotations augmenting the terminal reward and periodically distills them into a small set of natural-language strategic rules.
\recap requires no weight updates and learns purely from played episodes, substantially improving over both the base LLM and a case-based-reasoning baseline on \ttt with \gptnano against the \optimal opponent.

Our contributions are:
\begin{itemize}
  \item An empirical characterization of the gap in sequential reasoning in current LLMs on simple deterministic perfect-information two-player zero-sum games, showing it persists across models and providers, and is largely invariant to surface representation.
  \item \recap, an agentic framework that adapts case-based reasoning to sequential decision-making through per-move reflection and rule extraction, improving over a strong baseline without modifying the model weights.
\end{itemize}

\section{Related Work}\label{sec:related_work}

A growing body of work has examined how LLMs perform on reasoning tasks that require more than a single output step.
Chain-of-Thought prompting~\cite{cot} elicits intermediate reasoning steps and has become standard practice on multi-step problems, with extensions such as self-consistency~\cite{selfconsistency} improving robustness through majority voting over independent reasoning chains.
Tree of Thoughts~\cite{tot} reframes reasoning as a search problem, structuring intermediate thoughts into a tree, self-evaluating them and backtracking on dead ends.
Although closer to the explicit search used in sequential decision-making, it searches over the model's internal states rather than a shared external state.

A separate line of research targets environments with external state directly.
ReAct~\cite{react} interleaves reasoning and action steps in interactive environments, laying the foundations for agentic systems.
RAP~\cite{rap} prompts the LLM to act as both a world model and a value function, and plans over its own predictions using Monte Carlo Tree Search.
Code World Models~\cite{cwm} have the LLM synthesize executable code that represents the environment, then use a dedicated planning algorithm on the resulting world model.
These methods use the LLM for environment representation or prediction, but unlike our method, the planning is performed by an external algorithm.

A complementary approach is to keep the LLM as the primary reasoner and provide it with structured context drawn from past experience.
Memento~\cite{memento} introduces a case-based reasoning architecture in which an agent consults a growing \textit{case bank} of past task-solving trajectories, retrieving relevant cases at decision time.
ExpeL~\cite{expel} extracts natural-language insights from successful and failed trajectories and supplies them as additional context for new tasks.
Reflexion~\cite{reflexion} prompts the agent to reflect on its own failures and uses the resulting reflections as guidance on subsequent attempts at the same task.
These methods establish that experience memory and self-reflection can yield consistent gains without modifying model weights, but they were developed primarily for stationary question-answering and tool-use tasks, where the environment does not respond adversarially and trajectories are short enough that storing them as a single unit, labeled by the final outcome, gives a useful retrieval signal.
\recap adapts this family of methods to sequential play, where trajectories are longer, the binary win/loss reward arrives only at the end, a signal too coarse to tell strong moves from mistakes within a trajectory and drive learning.

A third line of work focuses on the infrastructure for evaluating LLMs on games.
OpenSpiel~\cite{openspiel} provides a comprehensive library of game implementations and reference algorithms, and is widely used as the standard engine for game-theoretic experiments.
GTBench~\cite{gtbench}, building on OpenSpiel, introduces a benchmark suite for evaluating LLMs on a range of (im)perfect-information, deterministic and stochastic games, and reports substantial gaps to optimal play, including on \ttt, \nim, and \cfour.
\omniplay extends GTBench into a modular evaluation framework that supports arbitrary players, games, and obfuscations through a small set of adapter interfaces, and serves as the platform for evaluating \recap; for differences from the concurrent OpenSpiel~2.0~\cite{openspiel2}, see \refsupp{app:openspiel}.

\section{The Sequential Reasoning Gap}\label{sec:evals}
In this section we diagnose the sequential reasoning gap by evaluating LLMs on deterministic perfect-information two-player zero-sum games.
As argued in \refsection{sec:intro}, such games provide a controlled setting for this purpose: the rules are well-defined, actions and final outcomes allow for unambiguous ground-truth evaluation, and the optimal policy is computable exactly for games of modest size.
We show that current \sota LLMs remain unreliable even in simple domains, then probe whether this unreliability is better explained by recall of memorized strategies or by flawed reasoning about the current state.
The gap to optimal play persists across models and opponents and is largely invariant to representation, motivating the method introduced in \refsection{sec:recap}.

\subsection{\omniplay}\label{sec:evals:omniplay}
To evaluate a broad spectrum of players, ranging from random and algorithmic baselines to LLM-based players, on arbitrary games, we developed \omniplay\footnote{GitHub repository: \urloranon{https://github.com/radajakub/plybench}}, shown in \reffigure{fig:evals:omniplay}.

Following GTBench~\cite{gtbench}, \omniplay uses OpenSpiel~\cite{openspiel} as the underlying game engine and introduces two components: the \textbf{Interface Adapter}, which translates OpenSpiel's observations and actions into a unified representation for all types of players; and the \textbf{LLM Adapter}, which transforms this representation into a text-based format suitable for LLMs.
New games and players are integrated by implementing the interfaces defined by these adapters, regardless of their internal complexity.
\omniplay then handles game execution, player orchestration, parallelization, and result aggregation and analysis.

\subsection{Evaluation Setting}\label{sec:eval:setting}
We consider three perfect-information two-player zero-sum games, each posing different challenges.
\textbf{\Ttt} features a small game tree and short-horizon planning, with draw as the optimal outcome, and thus tests the player's ability to avoid losing rather than to win.
\textbf{\nim} also has a small game tree, but requires a less obvious strategy based on the Nim-sum of the piles, and tests the player's ability to force a win when starting in a winning position.
\textbf{\cfour} has a much larger game tree, requiring long-horizon planning and the ability to anticipate and create multiple simultaneous threats.

To measure the game-playing ability of the players, we use three opponents of increasing strength.
The \textbf{Random} opponent chooses actions uniformly at random.
The \textbf{Monte Carlo Tree Search (MCTS)} opponent is strong but exploitable. 
Since MCTS estimates state values from random rollouts, a bounded simulation budget $n$ leaves a non-trivial probability of suboptimal play.
We use $n=1000$ for \ttt and \nim, and $n=5000$ for \cfour, large enough for strong play but allowing for occasional mistakes.
Other hyperparameters are set to the OpenSpiel defaults.
The \textbf{\optimal} opponent runs the \textit{\minimax} algorithm, breaking ties uniformly.

We evaluate \sota LLMs from OpenAI~\cite{gpt5card} and Google~\cite{gemini}, together with selected open-source models~\cite{qwen3,glm5}, grouped into three tiers (see \reftable{tab:eval:model_groups}) by expected capability, as reflected in public benchmarks~\cite{modelcomparison} rather than parameter count.
Unless stated otherwise, the models are queried with reasoning effort set to \textit{high}, if such a setting is exposed by the provider; all other hyperparameters use their default values.

\begin{table}[ht]
  \small
  \centering
  \begin{tabularx}{\columnwidth}{@{}lX@{}}
    \toprule
    Group & Models \\
    \midrule
    \efficient & \gptnano, \geminifl, \oss \\
    \intermediate & \gptmini, \geminif, \qwen \\
    \frontier & \gpt, \gemini, \glm \\
    \bottomrule
  \end{tabularx}
  \caption{LLMs grouped by expected capabilities into tiers.}
  \label{tab:eval:model_groups}
\end{table}

Under optimal play, \ttt is a draw for both players, whereas \cfour is a first-player win and \nim (from the selected initial position $\{1,3,5,7\}$) is a first-player loss.
For each player-opponent pair, we run $100$ games, switching sides after half of the games to control for the first or second player advantage.
For \ttt and \nim, we report per-move optimality rates, where a move is considered optimal if it could have been chosen by \minimax.
States in which all legal moves are optimal are excluded because they provide no discriminative signal.
For \cfour, where the game tree is too large to compute exact optimality easily, we instead report win rate and average number of player moves in the game.

\subsection{Failures in Simple Domains}\label{sec:evals:simple_domains}
We first evaluate the LLMs under the standard game rules.
We found the \frontier tier to play \nim optimally and \ttt near-optimally in this setting, so we restrict the analysis of these two games to the \efficient and \intermediate tiers.
The \frontier tier is evaluated below on \cfour, where optimal play is substantially more challenging.

\begin{table}[ht]
  \small
  \centering
  \begin{tabular}{lcc}
    \toprule
    Model & \multicolumn{1}{c}{\ttt} & \multicolumn{1}{c}{\nim} \\
    \midrule
    \gptnanoshort & \rci{$.880^{\dagger}$}{.828}{.918} & \rci{.955}{.904}{.979} \\
    \geminiflshort & \rci{.956}{.918}{.977} & \rci{1.00}{.976}{1.00} \\
    \ossshort & \rci{$.901^{\dagger}$}{.853}{.935} & \rci{.980}{.943}{.993} \\
    \midrule
    \gptminishort & \rci{.922}{.878}{.952} & \rci{.993}{.959}{.999} \\
    \geminifshort & \rci{.995}{.974}{.999} & \rci{1.00}{.974}{1.00} \\
    \qwenshort & \rci{.949}{.909}{.972} & \rci{1.00}{.974}{1.00} \\
    \bottomrule
  \end{tabular}
  \caption{Per-move optimality rates with \wilson for \efficient- and \intermediate-tier LLMs on \ttt and \nim against the \optimal opponent. Each model plays $100$ games, switching sides halfway through. $^{\dagger}$\gptnanoshort and \ossshort failed to return a legal action in $2$ and $8$ moves, respectively.}
  \label{tab:eval:simple_domains:optimality}
\end{table}

\reftable{tab:eval:simple_domains:optimality} shows that no model from the \efficient or \intermediate tier reaches optimal play on \ttt, even with high reasoning effort, and several models occasionally fail to produce a legal action.
Apart from \geminif, which makes roughly one mistake per $200$ moves, the remaining models make four to twelve mistakes per $100$ moves, a substantial error rate given the small game tree.
On \nim, the picture is mixed: several models reach optimal play, but the smaller models still miss a non-trivial fraction of moves.

The situation is significantly worse for the \frontier tier on \cfour.
\reftable{tab:eval:simple_domains:cfour_moves} reports win rates and the average number of moves made by the LLM player against the \random and \mcts opponents.
All three models defeat the \random opponent in every game, indicating that they can follow the rules and exploit unstructured play.
However, given the low probability of strong moves under random play, this is only a weak indicator of strategic competence.

\begin{table}[ht]
  \small
  \centering
  \setlength{\tabcolsep}{5pt}
  \begin{tabular}{lcccc}
    \toprule
    \multirow{2}[3]{*}{Model} & \multicolumn{2}{c}{\random} & \multicolumn{2}{c}{\mcts} \\
    \cmidrule(lr){2-3}\cmidrule(lr){4-5}
    & \multicolumn{1}{c}{Win rate} & \multicolumn{1}{c}{Moves} & \multicolumn{1}{c}{Win rate} & \multicolumn{1}{c}{\makecell{Moves}} \\
    \midrule
    \gptshort & \rci{1.00}{0.72}{1.00} & \rci{4.90}{4.04}{5.76} & \rci{0.00}{0.00}{0.28} & \rci{10.90}{6.93}{14.87} \\
    \geminishort & \rci{1.00}{0.72}{1.00} & \rci{5.40}{4.43}{6.37} & \rci{0.00}{0.00}{0.28} & \rci{13.90}{10.97}{16.83} \\
    \glmshort & \rci{1.00}{0.72}{1.00} & \rci{5.90}{4.23}{7.57} & \rci{0.00}{0.00}{0.28} & \rci{6.20}{3.70}{8.70} \\
    \bottomrule
  \end{tabular}
  \caption{Win rate and average number of moves made by the LLM player before the game ended on \cfour against the \random and \mcts opponents. Win rates are reported with \wilson, while the number of LLM moves is reported with \student. Each model plays $10$ games against each opponent, switching sides halfway through.}
  \label{tab:eval:simple_domains:cfour_moves}
\end{table}

Against the stronger but imperfect \mcts opponent, no \frontier model won a single game, and all were defeated rapidly.
Whereas bilateral perfect play in \cfour extends to move $41$, a first-player win, requiring at least $20$ moves per player, the \frontier models made only $10.3$ moves on average before losing.
This indicates that failure arises already in the opening phase rather than in tactically complex middlegames.

\subsection{Obfuscations}\label{sec:evals:obfuscations}
We have seen models fail even on simple games such as \ttt and \nim.
Both games are well known and likely appear frequently in pretraining data, including examples of good and bad play from many positions.
A natural alternative explanation is therefore that the observed errors reflect imperfect recall of familiar positions or strategies, rather than flawed reasoning about the current state.

\begin{figure}[ht]
  \centering
  \resizebox{0.85\linewidth}{!}{%
    \begin{tikzpicture}[
  font=\bfseries\sffamily,
  >=Latex,
  line cap=round,
  line join=round,
  gridCell/.style={
    draw=gray!65,
    line width=\msLineWidth,
    minimum width=11mm,
    minimum height=11mm,
    inner sep=0pt,
    align=center,
    fill=white,
    anchor=north west
  },
  rightGridCell/.style={
    gridCell,
    fill=\rightGridBG
  },
  sumCell/.style={
    draw=black,
    line width=\msThickLineWidth,
    rounded corners=\msItemCorner,
    minimum width=9mm,
    minimum height=9mm,
    inner sep=0pt,
    align=center,
    fill=white,
    anchor=north west
  },
  thickArrow/.style={->, line width=2.4pt},
  thinArrow/.style={->, line width=1.2pt}
]

\coordinate (left_tl) at (0,0);
\pgfmathsetlengthmacro{\cellSize}{11mm}
\pgfmathsetlengthmacro{\gap}{22mm}
\pgfmathsetlengthmacro{\diagDist}{7mm}
\pgfmathsetlengthmacro{\sumGap}{\diagDist/sqrt(2)}
\pgfmathsetlengthmacro{\sumSize}{9mm}

\SetLeftMark{0}{0}{\MarkX}
\SetRightCellBG{0}{0}{red!15}
\SetLeftMark{1}{1}{\MarkX}
\SetRightCellBG{1}{1}{red!15}
\SetLeftMark{0}{1}{\MarkX}
\SetRightCellBG{0}{1}{red!15}
\SetLeftMark{0}{2}{\MarkX}
\SetRightCellBG{0}{2}{red!15}
\SetTotalBG{rowSum0}{green!15}
\SetLeftMark{2}{2}{\MarkO}
\SetRightCellBG{2}{2}{blue!15}
\SetLeftMark{2}{0}{\MarkO}
\SetRightCellBG{2}{0}{blue!15}
\SetLeftMark{1}{0}{\MarkO}
\SetRightCellBG{1}{0}{blue!15}

\node[gridCell] (L00) at ($(left_tl)+(0*\cellSize,0*\cellSize)$) {\leftMark{0}{0}};
\node[gridCell] (L01) at ($(left_tl)+(1*\cellSize,0*\cellSize)$) {\leftMark{0}{1}};
\node[gridCell] (L02) at ($(left_tl)+(2*\cellSize,0*\cellSize)$) {\leftMark{0}{2}};
\node[gridCell] (L10) at ($(left_tl)+(0*\cellSize,-1*\cellSize)$) {\leftMark{1}{0}};
\node[gridCell] (L11) at ($(left_tl)+(1*\cellSize,-1*\cellSize)$) {\leftMark{1}{1}};
\node[gridCell] (L12) at ($(left_tl)+(2*\cellSize,-1*\cellSize)$) {\leftMark{1}{2}};
\node[gridCell] (L20) at ($(left_tl)+(0*\cellSize,-2*\cellSize)$) {\leftMark{2}{0}};
\node[gridCell] (L21) at ($(left_tl)+(1*\cellSize,-2*\cellSize)$) {\leftMark{2}{1}};
\node[gridCell] (L22) at ($(left_tl)+(2*\cellSize,-2*\cellSize)$) {\leftMark{2}{2}};

\coordinate (right_tl) at ($(left_tl)+(3*\cellSize+\gap,0)$);
\node[rightGridCell, fill=\rightCellBG{0}{0}] (R00) at ($(right_tl)+(0*\cellSize,0*\cellSize)$) {$2$};
\node[rightGridCell, fill=\rightCellBG{0}{1}] (R01) at ($(right_tl)+(1*\cellSize,0*\cellSize)$) {$7$};
\node[rightGridCell, fill=\rightCellBG{0}{2}] (R02) at ($(right_tl)+(2*\cellSize,0*\cellSize)$) {$6$};
\node[rightGridCell, fill=\rightCellBG{1}{0}] (R10) at ($(right_tl)+(0*\cellSize,-1*\cellSize)$) {$9$};
\node[rightGridCell, fill=\rightCellBG{1}{1}] (R11) at ($(right_tl)+(1*\cellSize,-1*\cellSize)$) {$5$};
\node[rightGridCell, fill=\rightCellBG{1}{2}] (R12) at ($(right_tl)+(2*\cellSize,-1*\cellSize)$) {$1$};
\node[rightGridCell, fill=\rightCellBG{2}{0}] (R20) at ($(right_tl)+(0*\cellSize,-2*\cellSize)$) {$4$};
\node[rightGridCell, fill=\rightCellBG{2}{1}] (R21) at ($(right_tl)+(1*\cellSize,-2*\cellSize)$) {$3$};
\node[rightGridCell, fill=\rightCellBG{2}{2}] (R22) at ($(right_tl)+(2*\cellSize,-2*\cellSize)$) {$8$};

\draw[thickArrow] ($(L12.east)+(3mm,0)$) -- ($(R10.west)+(-3mm,0)$);

\node[sumCell, fill=\totalBG{rowSum0}] (rowSum0) at
  ($(right_tl)+(3*\cellSize+\sumGap,-0*\cellSize-0.5*\cellSize+0.5*\sumSize)$) {$15$};
\draw[thinArrow] (R02.east) -- (rowSum0.west);
\node[sumCell, fill=\totalBG{rowSum1}] (rowSum1) at
  ($(right_tl)+(3*\cellSize+\sumGap,-1*\cellSize-0.5*\cellSize+0.5*\sumSize)$) {$15$};
\draw[thinArrow] (R12.east) -- (rowSum1.west);
\node[sumCell, fill=\totalBG{rowSum2}] (rowSum2) at
  ($(right_tl)+(3*\cellSize+\sumGap,-2*\cellSize-0.5*\cellSize+0.5*\sumSize)$) {$15$};
\draw[thinArrow] (R22.east) -- (rowSum2.west);

\node[sumCell, fill=\totalBG{colSum0}] (colSum0) at
  ($(right_tl)+(0*\cellSize+0.5*\cellSize-0.5*\sumSize,-3*\cellSize-\sumGap)$) {$15$};
\draw[thinArrow] (R20.south) -- (colSum0.north);
\node[sumCell, fill=\totalBG{colSum1}] (colSum1) at
  ($(right_tl)+(1*\cellSize+0.5*\cellSize-0.5*\sumSize,-3*\cellSize-\sumGap)$) {$15$};
\draw[thinArrow] (R21.south) -- (colSum1.north);
\node[sumCell, fill=\totalBG{colSum2}] (colSum2) at
  ($(right_tl)+(2*\cellSize+0.5*\cellSize-0.5*\sumSize,-3*\cellSize-\sumGap)$) {$15$};
\draw[thinArrow] (R22.south) -- (colSum2.north);

\node[sumCell, fill=\totalBG{diagSum}] (diagSum) at
  ($(right_tl)+(3*\cellSize+\sumGap,-3*\cellSize-\sumGap)$) {$15$};
\draw[thinArrow] (R22.south east) -- (diagSum.north west);

\node[sumCell, fill=\totalBG{leftBottomSum}] (leftBottomSum) at
  ($(right_tl)+(-\sumGap-\sumSize,-3*\cellSize-\sumGap)$) {$15$};
\draw[thinArrow] (R20.south west) -- (leftBottomSum.north east);

\path[use as bounding box]
  ($(L00.north west)+(-2mm,2mm)$)
  rectangle
  ($(diagSum.south east)+(2mm,-2mm)$);
\end{tikzpicture}
  }
  \caption{A Magic Square obfuscation of \ttt. The original game board is mapped to a Magic Square grid, where each cell is assigned a number from $1$ to $9$ such that the sum of each row, column and diagonal is $15$. Players alternately take numbers from this set; the first one to collect three that sum to $15$ wins.}
  \label{fig:evals:obfuscations:magic_square}
\end{figure}
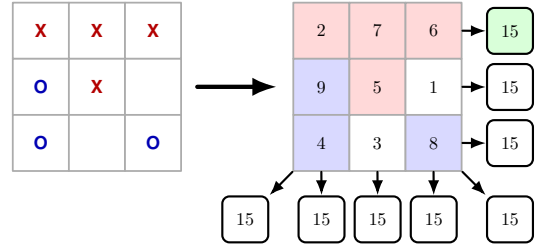

To probe the role of memorized strategies, we introduce alternative rule formulations that preserve the underlying game tree while changing its surface representation.
Each such \textit{obfuscation} is structurally identical to the original game, but the state observations and available actions, both expressed in natural language, are different.

We focus here on the \ttt obfuscations, which provide a compact representative case; \nim is evaluated with analogous surface transformations and presented in \refsupp{app:obfuscations:nim}.
Since these formulations are unlikely to appear in pretraining data, a model relying on memorized positions or strategies tied to the original representation should play noticeably worse on them, whereas a model that reasons about the current state should be largely unaffected.

\textbf{\hidlong} (\hid) uses the same rules as \ttt except that the markers \texttt{X} and \texttt{O} are replaced with \texttt{B} and \texttt{W}, and the game name is not mentioned.
\textbf{\msqlong} (\msq) is played on a set of numbers $\{1, \ldots, 9\}$.
Players alternately claim a number that has not yet been taken, and the first to hold three numbers summing to $15$ wins.
This is isomorphic to \ttt: the nine numbers can be arranged in a $3\times3$ grid where every row, column, and diagonal sums to $15$.
An example configuration is shown in \reffigure{fig:evals:obfuscations:magic_square}.
\textbf{\msqaddlong} (\msqadd) is identical to \msqlong, but every cell value is incremented by $5$ to obscure the standard pattern so that the winning triples now sum to $30$.
\textbf{\storylong} (\story) encodes the \msq obfuscation as a story.
The players try to cross a river of width $15$, each aiming to do so in three jumps.
On their turn, a player claims a jump length from $\{1, \ldots, 9\}$; each jump length may be claimed only once.
A player wins if three of their claimed lengths sum to $15$, which is a crossing that neither overshoots nor undershoots.

\begin{table}[ht]
  \centering
  \small
  \setlength{\tabcolsep}{2pt}
  \begin{tabular}{lccccc}
    \toprule
    Model & \tttshort & \hid & \msq & \msqadd & \story \\
    \midrule
    \gptnanoshort & \rci{.880}{.828}{.918} & \rci{.871}{.817}{.911} & \rci{.888}{.837}{.925} & \rci{.785}{.718}{.840} & \rci{.844}{.786}{.888} \\
    \geminiflshort & \rci{.956}{.918}{.977} & \rci{.950}{.910}{.972} & \rci{.991}{.966}{.997} & \rci{.991}{.966}{.997} & \rci{.703}{.635}{.763} \\
    \ossshort & \rci{.901}{.853}{.935} & \rci{.887}{.837}{.924} & \rci{.901}{.853}{.935} & \rci{.779}{.716}{.832} & \rci{.835}{.777}{.880} \\
    \midrule
    \gptminishort & \rci{.922}{.878}{.952} & \rci{.923}{.878}{.952} & \rci{.967}{.934}{.984} & \rci{.971}{.939}{.987} & \rci{.953}{.916}{.974} \\
    \geminifshort & \rci{.995}{.974}{.999} & \rci{1.00}{.983}{1.00} & \rci{1.00}{.983}{1.00} & \rci{.995}{.974}{.999} & \rci{1.00}{.983}{1.00} \\
    \qwenshort & \rci{.949}{.909}{.972} & \rci{.956}{.918}{.977} & \rci{1.00}{.982}{1.00} & \rci{.986}{.960}{.995} & \rci{.721}{.656}{.779} \\
    \bottomrule
  \end{tabular}
  \caption{Per-move optimality rates with \wilson for \efficient- and \intermediate-tier LLMs on \ttt and its obfuscations against the \optimal opponent. Each model plays $100$ games per rule formulation, switching sides halfway through.}
  \label{tab:eval:obfuscations:ttt_results}
\end{table}

The evaluation setting mirrors that of \refsection{sec:evals:simple_domains}.
As shown in \reftable{tab:eval:obfuscations:ttt_results}, the hardest obfuscations do induce declines for some models, specifically \gptnanoshort and \ossshort on \msqadd, and \geminiflshort and \qwenshort on \story.
These declines are idiosyncratic, however, with no failure mode generalizing across both models and obfuscations.
This suggests that the models are not relying solely on memorized \ttt positions, but can often adapt their play to semantically equivalent rule formulations.

To further inspect the reasoning process, we examine reasoning-trace summaries and compute the proportion that explicitly names the underlying game (\reftable{tab:eval:obfuscations:ttt_recognition}); since a model may recognize the hidden structure without verbalizing it, this is a lower bound.
Models identify the original game in a substantial fraction of traces for the milder obfuscations, \hid and \msq, but recognition drops sharply on \msqadd and \story.
This suggests they often infer the hidden structure from the rules rather than merely match the original surface form.
Once the game is recognized, we cannot rule out the use of memorized positions or strategies.
If this flawed recall were the source of the residual sub-optimality, however, the gap to optimal play should widen when the model no longer recognizes the game.
Yet \geminiflshort recognizes \ttt in only a quarter of its \msqadd traces while its optimality stays near-perfect, so the gap there cannot be attributed to flawed recall only; on \story, recognition collapses even further while optimality falls far less steeply.
This points to decision-time reasoning rather than flawed memoization.

\begin{table}[ht]
  \centering
  \small
  \setlength{\tabcolsep}{3pt}
  \begin{tabular}{lcccc}
    \toprule
    Model & Hid & M.Sq & M.Sq +5 & Story \\
    \midrule
    \gptnanoshort & \rci{.830}{.791}{.863} & \rci{.931}{.902}{.951} & \rci{.424}{.377}{.472} & \rci{.482}{.435}{.530} \\
    \geminiflshort & \rci{.998}{.987}{1.00} & \rci{.922}{.893}{.943} & \rci{.256}{.218}{.298} & \rci{.189}{.154}{.231} \\
    \ossshort & \rci{.944}{.918}{.963} & \rci{.951}{.926}{.968} & \rci{.539}{.486}{.591} & \rci{.736}{.691}{.776} \\
    \midrule
    \gptminishort & \rci{.364}{.317}{.414} & \rci{.588}{.538}{.635} & \rci{.581}{.534}{.628} & \rci{.859}{.822}{.889} \\
    \geminifshort & \rci{.991}{.977}{.997} & \rci{.831}{.794}{.863} & \rci{.414}{.369}{.460} & \rci{.756}{.714}{.793} \\
    \qwenshort & \rci{1.00}{.991}{1.00} & \rci{.996}{.984}{.999} & \rci{.633}{.587}{.676} & \rci{.435}{.387}{.485} \\
    \bottomrule
  \end{tabular}
  \caption{Proportion of reasoning traces that explicitly identify the underlying game as \ttt under each obfuscation. Results are aggregated over $100$ games per model-obfuscation pair ($\sim400$ traces per cell), and reported with \wilson.}
  \label{tab:eval:obfuscations:ttt_recognition}
\end{table}

Although optimality rates vary little across obfuscations, the number of output tokens per move increases steadily with obfuscation complexity, with token statistics reported in \refsupp{app:costs:tokens}.
This suggests that models expend additional reasoning effort to interpret the harder formulations and map them to playable states.

Overall, this section shows that current \sota LLMs remain unreliable in simple sequential decision-making domains.
Models in the \efficient and \intermediate tiers fall short on \ttt and \nim, while the \frontier tier fails early on \cfour.
Performance remains largely invariant across surface-level obfuscations, and reasoning traces show that models frequently identify the obscured game from the rules alone.
Together, these findings argue against a pure surface-memorization explanation and point to a decision-time reasoning bottleneck.
Additional results are reported in \refthreesupps{app:evals}{app:obfuscations:nim}{app:costs}.

\begin{figure*}[!t]
  \centering
  \resizebox{0.95\linewidth}{!}{%
    \input{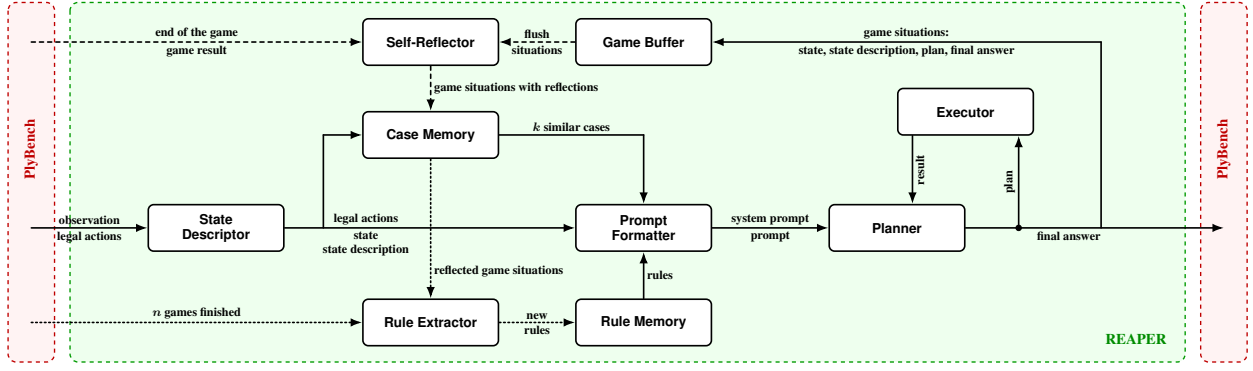}
  }
  \caption{Schema of the \recap architecture and its interaction with \omniplay.}
  \label{fig:memento:architecture}
\end{figure*}

\section{Reflective Case-Based Reasoning}\label{sec:recap}

\refsection{sec:evals} showed that LLMs remain unreliable in simple perfect-information games.
This unreliability is largely invariant to surface-level changes in the game rules, suggesting a decision-time reasoning bottleneck rather than faulty recall of memorized strategies.
A natural way to address this bottleneck without modifying model weights is to provide the model with structured context from past experience, as Memento~\cite{memento} and ExpeL~\cite{expel} do for question-answering and tool-use tasks.
Although these tasks may involve sequences of reasoning steps, they differ from the sequential decision-making setting studied here.
The environment does not evolve in response to the agent, and there is no adversary whose actions shape the outcome.

Adapting case-based reasoning to this setting raises challenges that the mentioned methods do not address.
The unit of experience is an individual move, rather than a completed task, yet reward is revealed only at the episode's end, so the terminal outcome must be distributed across the full trajectory.
We address these challenges in \recap, a reflective case-based reasoning framework that keeps the LLM as the primary reasoner while augmenting it with an experience memory designed for sequential play.
We first describe the progression from a direct Memento port to a working baseline and then introduce the additional components for \recap.

\subsection{From Static Case Memory to Sequential Play}\label{sec:recap:working_agent}

We begin with a minimal adaptation of Memento~\cite{memento} to sequential game play, which we call \original.
To solve a task, Memento uses an LLM planner-executor loop.
The planner generates either a plan, which is executed step-by-step, or a final answer that is returned to the environment.
Completed tasks are stored in case memory together with the latest plan and an outcome score produced by an LLM judge.
At decision time, similar cases are retrieved to condition the solution of future tasks.

To adapt this structure to sequential play, we add a \textbf{Game Buffer} that stores generated moves until the episode terminates.
Once the outcome is observed, the same reward is assigned to all buffered moves, and the resulting cases are written to memory.
Each move is a complete planner-executor task, so the natural unit of experience is a single move rather than a whole trajectory, which would bundle many tasks into one case.
Under full observability this is also sound for retrieval: a decision's quality depends on the current state, not the path to it.
Since game outcomes are determined exactly by the environment, no LLM judge is needed.
This minimal port suffices for environment interaction, but not for dependable play.
Its prompt-level formatting instructions and response parsing are brittle in settings that require precise action formats, leading to frequent failures.
Also, storing the same decision repeatedly lets the majority outcome for a given state dominate rarer but valid ones at retrieval time.

We therefore introduce a stronger engineered variant, denoted \baseline.
It preserves the planner-executor architecture and case memory, but makes inter-component communication robust through structured outputs, reducing parsing errors.
The final answer is generated as a free-form string to keep the method independent of game-specific action encoding.
Memory entries are deduplicated by state-action-reward triples, preventing domination by repeated identical cases.
Diversity of retrieved cases is improved by stratified sampling that balances positive and negative examples.

Although these changes make \baseline a usable game-playing agent, they do not address structural limitations of case-based reasoning in games.
The credit assignment is coarse: every move in a lost game receives negative credit, even if some were locally optimal, and every move in a won game receives positive credit, even if some were locally suboptimal.
A second limitation concerns decision-time retrieval of similar cases, which is limited to already observed states and actions with no mechanism to generalize to unseen states, leading to slow learning in large domains.

\subsection{\recap}\label{sec:recap:architecture}

\recap, shown in \reffigure{fig:memento:architecture}, retains the planner-executor architecture and a case memory, and adds two components: the \textbf{self-reflector} and the \textbf{rule extractor}, which address the two limitations identified above.
A third auxiliary component, the \textbf{state descriptor}, transforms observations into natural language before retrieval and is evaluated separately in \refsupp{app:recap:state_descriptor}.

The \textbf{self-reflector} targets the coarse credit assignment of \baseline.
Once an episode terminates, it analyzes the full trajectory and produces structured per-move reflections, instead of assigning a single terminal reward uniformly over every move.
The generated reflection scores a move along two independent axes: its local quality, whether the move was sound at the point it was played, and its outcome contribution, whether it helped, hurt, or was neutral to the final result.
This decomposition is the core of the finer-grained signal, recording a locally sound move in a lost game as good rather than uniformly penalizing it.
Reflections are stored alongside the terminal reward and are presented in the planner's prompt as concrete plays to emulate or mistakes to avoid.

Reflections only provide signal for the states actually encountered.
For large domains with many possible states, this can be a learning bottleneck, as it requires many interactions to explore the space sufficiently.
The \textbf{rule extractor} generalizes the reflections to guide the planner even in unseen states.
Periodically, after a fixed number of games, it consolidates the reflected cases accumulated since the last update, together with the current rule set, into a compact set of general strategic rules that are inserted into the planner's prompt.
Because a new set is generated each period, the rule memory is self-correcting, merging overlapping rules into stronger ones and removing rules contradicted by new evidence.
Together, the two components complement each other, with reflections attaching fine-grained credit to individual cases and rules distilling those into transferable principles.

\recap runs an online decision loop during play.
On each move, the current observation and legal actions retrieve the most semantically similar cases from case memory and the current rules from rule memory; these are combined into the planner's prompt, and the planner-executor loop runs until it returns an action, which is played and appended to the game buffer.
Then, the two post-game updates are performed: the self-reflector after each game and the rule extractor after a fixed number of games.
Full details of retrieval, the decision loop, and the update schedule are given in \refsupp{app:recap}.

\section{Experiments}\label{sec:experiments}

We ask whether reflective case-based memory improves sequential decision-making.
We test this on \ttt against the \optimal opponent: it admits exact move-level evaluation, yet \refsection{sec:evals} showed that \efficient models still make frequent mistakes on it.

\refsection{sec:evals} queried each model with \textit{high} reasoning effort to elicit its best single-call play; here we instead query \gptnanoshort at \textit{medium} effort.
This leaves the model more headroom to improve in.
At \textit{high} effort it already plays close to optimally, so gains from memory are small and hard to separate from noise; at \textit{medium} effort the wider gap to optimal play lets \recap's gains stand out and each component's contribution be distinguished more clearly.

\begin{figure}[ht]
  \centering
  \includegraphics[width=0.9\linewidth]{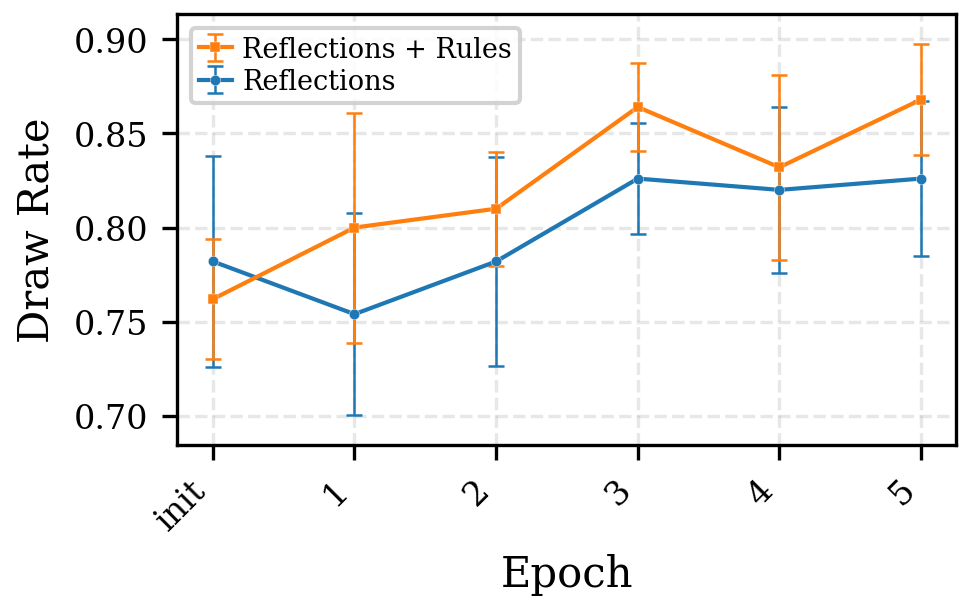}
  \caption{Ablation of individual \recap components using \gptnano with medium reasoning effort against the \optimal opponent on \ttt. Results are averaged over $10$ runs, error bars show \student.}
  \label{fig:experiments:ttt:ablation}
\end{figure}

\subsection{Evaluation Setting}\label{sec:experiments:setting}

Each run begins with 50 evaluation games on an empty memory, comparing the planner-executor architecture with the single-call reference.
Then it proceeds through 5 epochs of 20 learning games followed by 50 evaluation games.
During learning, the agent works as described in \refsection{sec:recap:architecture}: it buffers its decisions, generates reflections after each game, and writes the reflected cases to memory.
The rule extractor runs once at the end of each learning phase, consolidating the recent cases with the existing rules.
During evaluation, memory is frozen and no reflection or rule extraction occurs, so each evaluation phase measures the policy induced by the current memory snapshot.
We repeat every configuration over 10 independent runs.

We report draw rate as our primary metric, since it is the best achievable result in \ttt against the \optimal opponent.
We also report output tokens per move as a measure of inference cost and reasoning efficiency.

\begin{figure}[ht]
  \centering
  \begin{subfigure}[t]{0.9\linewidth}
    \centering
    \includegraphics[width=\linewidth]{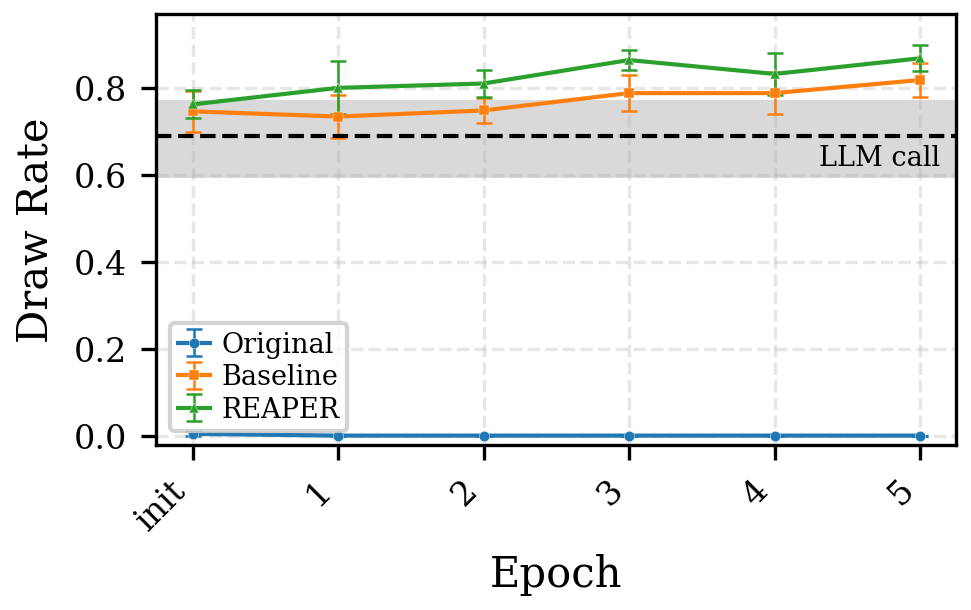}
    \caption{Draw rate}
    \label{fig:experiments:ttt:comparison:draw}
  \end{subfigure}

  \begin{subfigure}[t]{0.9\linewidth}
    \centering
    \includegraphics[width=\linewidth]{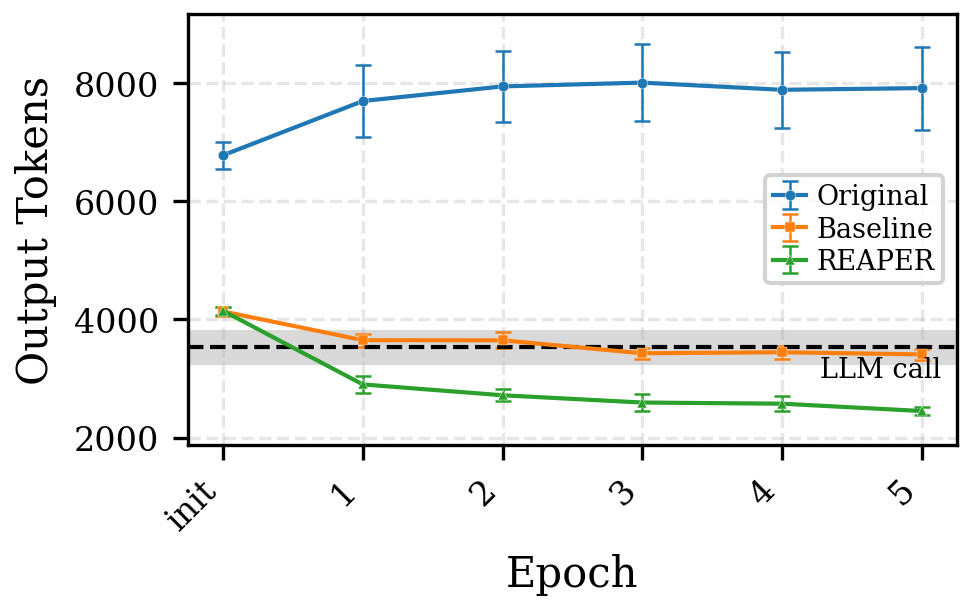}
    \caption{Output tokens, including reasoning tokens, per move.}
    \label{fig:experiments:ttt:comparison:tokens}
  \end{subfigure}
  \caption{Comparison of the \original port, the engineered \baseline, and the best \recap variant on \ttt with \gptnano (medium) against the \optimal opponent. Single-call reference from \refsection{sec:evals} is shown as a dashed line. Lines show means over $10$ independent runs, with error bars denoting \student.}
  \label{fig:experiments:ttt:comparison}
\end{figure}

\subsection{Ablation of \recap Components}\label{sec:experiments:ablation}

We ablate the two components of \recap.
Reflection is included in both variants, since per-move credit assignment is the core mechanism of \recap and the source from which rules are distilled; this gives two configurations, reflection alone and reflection with rules.
The state descriptor is an auxiliary component, evaluated separately in \refsupp{app:recap:state_descriptor}.

\reffigure{fig:experiments:ttt:ablation} shows that adding rule extraction improves over reflection alone.
The reflection-only variant reaches a mean draw rate of $0.826$, adding rules increases it to $0.868$.
Although the confidence intervals overlap, the improvement is significant under a one-sided Welch $t$-test ($p=0.039$).
An analogous, significant improvement occurs in per-move optimality rate, improving from $0.959$ to $0.970$ ($p=0.027$).

\subsection{Improvement over \original and \baseline}\label{sec:experiments:comparison}

We next compare \recap, in its configuration with reflection and rules, against the direct Memento port (\original) and the engineered \baseline.
We also include the single-call \gptnano player from \refsection{sec:evals} as a fixed reference.

\reffigure{fig:experiments:ttt:comparison} shows that \original fails to produce a usable game-playing agent.
Its draw rate remains at zero across all evaluation checkpoints, with most games ending in failure due to invalid moves or failed parsing.
In contrast, \baseline reaches a final mean draw rate of $0.818$, showing that strict planner-executor communication and memory deduplication can lead to viable play.

\recap improves further, reaching a final mean draw rate of $0.868$, a gain of $5$ percentage points over \baseline that is significant under a one-sided Welch $t$-test ($p=0.018$).
Since \recap and \baseline share the same underlying architecture, this gain isolates the contribution of post-game reflection and rule extraction.

\reffigure{fig:experiments:ttt:comparison} also reports output tokens per move.
\recap reduces token usage over training and ends below both \baseline and even the single-call reference.
The performance gain therefore does not come from longer reasoning traces, but from more useful decision-time context.

\section{Conclusion}\label{sec:conclusion}

We studied sequential decision-making in LLMs through perfect-information two-player zero-sum games, where both outcomes and individual moves can be evaluated against ground truth.
Across models, we found that even simple games expose a persistent decision-time reasoning gap, and that this gap is largely invariant to surface-level rule obfuscations.
We introduced \recap, a reflective case-based reasoning framework that adapts experience memory to sequential play.
Its central mechanism is per-move reflection that assigns credit under delayed rewards, complemented by rule extraction that generalizes this signal, and structured planner-executor communication that makes play reliable.
On \ttt with \gptnano, \recap improves over both a direct Memento port and a strong engineered baseline, while reducing output-token usage.
These results suggest that external experience memory can improve LLM agents without modifying model weights, but they also leave open how well the approach scales to larger games, longer horizons, and partial observability.

\appendix

\section*{Acknowledgments}

This research is supported by the Czech Science Foundation GA25-18353S and the CTU in Prague grant SGS23/184/OHK3/3T/13.
The access to the computational infrastructure of the OP VVV funded project CZ.02.1.01/0.0/0.0/16\_019/0000765 is also acknowledged.

\bibliographystyle{named}
\bibliography{ijcai26}

\clearpage

\section{Additional Standard Evaluation Results}\label{app:evals}

This supplementary section complements \refsection{sec:evals} by reporting full game-level outcomes on the standard versions of \ttt and \nim.
The main text reports per-move optimality against the \optimal opponent for the \efficient and \intermediate tiers, since the \frontier tier plays these two games near-optimally.
Here, we instead report wins, draws, losses, and failures against the \random, \mcts, and \optimal opponents, and we extend the analysis to the \frontier tier.

The evaluation setting matches \refsection{sec:eval:setting}: each model-opponent pair plays $100$ games with sides switched halfway to control for the first or second player advantage, and the \mcts opponent uses the same simulation budget $n=1000$.
The two games differ in their optimal outcome.
Under optimal play \ttt is a draw, whereas \nim from the initial position $\{1,3,5,7\}$ is a second-player win.
A failure denotes an illegal or ill-formed action that \omniplay could not parse.
We report raw counts of wins, draws, losses, and failures rather than confidence intervals to keep the four-way decomposition legible.

\begin{table}[h]
  \small
  \centering
  \setlength{\tabcolsep}{3pt}
  \begin{tabular}{l @{}p{9pt}@{} cccc @{}p{9pt}@{} cccc @{}p{9pt}@{} cccc}
    \toprule
    \multirow{2}{*}{Model} && \multicolumn{4}{c}{Random} & & \multicolumn{4}{c}{MCTS} & & \multicolumn{4}{c}{Optimal} \\
    \cmidrule(l{0pt}r{0pt}){3-6}\cmidrule(l{0pt}r{0pt}){8-11}\cmidrule(l{0pt}r{0pt}){13-16}
    && W & D & L & F & & W & D & L & F & & W & D & L & F \\
    \midrule
    \gptnanoshort && 89 & 8 & 0 & 3 && 0 & 75 & 23 & 2 && 0 & 76 & 22 & 2 \\
    \geminiflshort && 92 & 8 & 0 & 0 && 0 & 90 & 10 & 0 && 0 & 91 & 9 & 0 \\
    \ossshort && 83 & 9 & 1 & 7 && 0 & 68 & 9 & 23 && 0 & 80 & 12 & 8 \\
    \midrule
    \gptminishort && 91 & 8 & 1 & 0 && 0 & 93 & 7 & 0 && 0 & 84 & 16 & 0 \\
    \geminifshort && 96 & 4 & 0 & 0 && 1 & 98 & 1 & 0 && 0 & 99 & 1 & 0 \\
    \qwenshort && 93 & 7 & 0 & 0 && 0 & 93 & 7 & 0 && 0 & 90 & 10 & 0 \\
    \midrule
    \gptshort && 95 & 4 & 1 & 0 && 1 & 96 & 3 & 0 && 0 & 96 & 4 & 0 \\
    \geminishort && 93 & 7 & 0 & 0 && 0 & 100 & 0 & 0 && 0 & 100 & 0 & 0 \\
    \glmshort && 92 & 8 & 0 & 0 && 0 & 98 & 2 & 0 && 0 & 91 & 9 & 0 \\
    \bottomrule
  \end{tabular}
  \caption{Wins, draws, losses, and failures out of $100$ games for all models on standard \ttt.}
  \label{tab:app:evals:ttt}
\end{table}

For most models, the \mcts and \optimal columns are similar: the near-optimal \frontier tier draws almost every game against both opponents (\gemini draws all $100$), and the remaining models show no systematic advantage for either opponent.
This suggests that outcomes on \ttt are not determined solely by opponent strength.
The clear exception is \oss, which does worse against the weaker \mcts opponent ($68$ draws, $23$ failures) than against \optimal ($80$ draws, $8$ failures).
Since \mcts is the weaker opponent, this inversion cannot be explained by opponent strength alone.
A plausible explanation is that \optimal produces canonical optimal-play trajectories, whereas \mcts occasionally plays suboptimally and steers the game into less common states; if such states are underrepresented in training data, they are harder for models that partly rely on learned positional priors.
Notably, most of \oss's degradation shows up as failures ($23$ versus $8$), indicating that these off-distribution states hurt not only strategic quality but also its ability to produce well-formed actions.

\begin{table}[ht]
  \small
  \centering
  \setlength{\tabcolsep}{3pt}
  \begin{tabular}{l @{}p{9pt}@{} cccc @{}p{9pt}@{} cccc @{}p{9pt}@{} cccc}
    \toprule
    \multirow{2}{*}{Model} && \multicolumn{4}{c}{Random} & & \multicolumn{4}{c}{MCTS} & & \multicolumn{4}{c}{Optimal} \\
    \cmidrule(l{0pt}r{0pt}){3-6}\cmidrule(l{0pt}r{0pt}){8-11}\cmidrule(l{0pt}r{0pt}){13-16}
    && W & D & L & F & & W & D & L & F & & W & D & L & F \\
    \midrule
    \gptnanoshort && 96 & 0 & 4 & 0 && 72 & 0 & 28 & 0 && 44 & 0 & 56 & 0 \\
    \geminiflshort && 100 & 0 & 0 & 0 && 95 & 0 & 5 & 0 && 50 & 0 & 50 & 0 \\
    \ossshort && 95 & 0 & 5 & 0 && 84 & 0 & 16 & 0 && 47 & 0 & 53 & 0 \\
    \midrule
    \gptminishort && 88 & 0 & 12 & 0 && 68 & 0 & 32 & 0 && 49 & 0 & 51 & 0 \\
    \geminifshort && 100 & 0 & 0 & 0 && 98 & 0 & 2 & 0 && 50 & 0 & 50 & 0 \\
    \qwenshort && 98 & 0 & 2 & 0 && 81 & 0 & 18 & 1 && 50 & 0 & 50 & 0 \\
    \midrule
    \gptshort && 100 & 0 & 0 & 0 && 86 & 0 & 14 & 0 && 50 & 0 & 50 & 0 \\
    \geminishort && 100 & 0 & 0 & 0 && 100 & 0 & 0 & 0 && 50 & 0 & 50 & 0 \\
    \glmshort && 99 & 0 & 1 & 0 && 94 & 0 & 6 & 0 && 50 & 0 & 50 & 0 \\
    \bottomrule
  \end{tabular}
  \caption{Wins, draws, losses, and failures out of $100$ games for all models on standard \nim.}
  \label{tab:app:evals:nim}
\end{table}

Unlike \ttt, \nim admits a compact symbolic strategy: optimal play follows from the Nim-sum of the piles rather than from reasoning over spatial threats, blocks, and forks.
This suggests that, for current models, applying a known symbolic evaluation might be easier than robust positional reasoning.
This likely explains why models handle \nim more reliably than \ttt.
It also shows that the chosen simulation budget is low enough that \mcts makes frequent mistakes in \nim, where the action space is larger than in \ttt, to allow stronger models to win nearly every game.

On \nim, \random is almost saturated for every model, whereas \mcts is the discriminating opponent, with win rates ranging from $68$ to a perfect $100$.
That the stronger models beat \mcts in nearly every game, including the half in which \mcts moves second and could force a win, indicates that at the chosen simulation budget ($n=1000$) \mcts frequently plays suboptimally in \nim.
The stronger models are able to exploit these mistakes more effectively than the weaker models.
We don't really observe this on \ttt, which might be attributed to the smaller action space of \ttt and hence more precise value estimates from the simulations.

The \optimal opponent is only challenging for the \efficient tier; the two stronger tiers solve it near-optimally.
Failures are almost absent on \nim, in contrast to \ttt, suggesting that production of invalid actions and formatting errors depend strongly on the game representation and action interface rather than only on model capability.

\clearpage
\section{Additional Obfuscation Results}\label{app:obfuscations:nim}
This section complements \refsection{sec:evals:obfuscations} in two ways.
First, we extend the \ttt obfuscation analysis to the \frontier tier in \refsupp{app:obfuscations:ttt:results}, which the main text omitted because it already plays the standard games near-optimally.
Second, we introduce analogous obfuscations for \nim in \refsupp{app:obfuscations:nim:rules} and report their results in \refsection{app:obfuscations:nim:results}, testing whether the invariance observed on \ttt also holds in a game whose optimal strategy rests on a compact symbolic state value.

As in \refsection{sec:evals:obfuscations}, all results are measured against the \optimal opponent using per-move optimality, excluding states where all legal moves are optimal, and we additionally report the proportion of reasoning traces that explicitly identify the underlying game, mirroring the recognition analysis in \refsection{sec:evals:obfuscations}.
It should be noted again that a model may recognize the game structure without explicitly verbalizing it, hence these recognition rates are lower bounds.

\subsection{\ttt Obfuscation Results}\label{app:obfuscations:ttt:results}

The \frontier tier confirms the representation invariance observed for \efficient and \intermediate tier models on \ttt in the main text.
Per-move optimality remains high and nearly flat across all obfuscations, as shown in \reftable{tab:app:obfuscations:ttt_results}.
\gemini plays optimally throughout, while \gpt and \glm stay close to optimal with no systematic decline on the harder formulations.

\begin{table}[ht]
  \centering
  \small
  \setlength{\tabcolsep}{2pt}
  \begin{tabular}{lccccc}
    \toprule
    Model & \tttshort & \hid & \msq & \msqadd & \story \\
    \midrule
    \gptnanoshort & \rci{.880}{.828}{.918} & \rci{.871}{.817}{.911} & \rci{.888}{.837}{.925} & \rci{.785}{.718}{.840} & \rci{.844}{.786}{.888} \\
    \geminiflshort & \rci{.956}{.918}{.977} & \rci{.950}{.910}{.972} & \rci{.991}{.966}{.997} & \rci{.991}{.966}{.997} & \rci{.703}{.635}{.763} \\
    \ossshort & \rci{.901}{.853}{.935} & \rci{.887}{.837}{.924} & \rci{.901}{.853}{.935} & \rci{.779}{.716}{.832} & \rci{.835}{.777}{.880} \\
    \midrule
    \gptminishort & \rci{.922}{.878}{.952} & \rci{.923}{.878}{.952} & \rci{.967}{.934}{.984} & \rci{.971}{.939}{.987} & \rci{.953}{.916}{.974} \\
    \geminifshort & \rci{.995}{.974}{.999} & \rci{1.00}{.983}{1.00} & \rci{1.00}{.983}{1.00} & \rci{.995}{.974}{.999} & \rci{1.00}{.983}{1.00} \\
    \qwenshort & \rci{.949}{.909}{.972} & \rci{.956}{.918}{.977} & \rci{1.00}{.982}{1.00} & \rci{.986}{.960}{.995} & \rci{.721}{.656}{.779} \\
    \midrule
    \gptshort & \rci{.981}{.952}{.993} & \rci{.971}{.939}{.987} & \rci{1.00}{.983}{1.00} & \rci{.995}{.975}{.999} & \rci{.991}{.967}{.997} \\
    \geminishort & \rci{1.00}{.983}{1.00} & \rci{1.00}{.983}{1.00} & \rci{1.00}{.983}{1.00} & \rci{1.00}{.983}{1.00} & \rci{1.00}{.983}{1.00} \\
    \glmshort & \rci{.957}{.920}{.977} & \rci{.982}{.954}{.993} & \rci{1.00}{.983}{1.00} & \rci{.937}{.892}{.963} & \rci{.959}{.924}{.978} \\
    \bottomrule
  \end{tabular}
  \caption{Per-move optimality rates with \wilson for all model tiers on \ttt and its obfuscations against the \optimal opponent. Each model plays $100$ games per rule formulation, switching sides halfway through.}
  \label{tab:app:obfuscations:ttt_results}
\end{table}

The \msq obfuscation is the least challenging across all model tiers, showing even better performance than the unobfuscated \ttt formulation.
This can be attributed to the \textit{Magic Square} being a well-known mathematical puzzle that the models have likely encountered before.
Applying a fixed offset in \msqadd pushes the puzzle further out of distribution: recognition falls sharply for nearly every model, yet optimality drops only for \gptnano, \oss, and \glm and stays essentially flat for the rest.
Competent play therefore does not require recognizing the original game, reinforcing the decision-time-reasoning conclusion of the main text.

\begin{table}[ht]
  \centering
  \small
  \setlength{\tabcolsep}{3pt}
  \begin{tabular}{lcccc}
    \toprule
    Model & Hid & M.Sq & M.Sq +5 & Story \\
    \midrule
    \gptnanoshort & \rci{.830}{.791}{.863} & \rci{.931}{.902}{.951} & \rci{.424}{.377}{.472} & \rci{.482}{.435}{.530} \\
    \geminiflshort & \rci{.998}{.987}{1.00} & \rci{.922}{.893}{.943} & \rci{.256}{.218}{.298} & \rci{.189}{.154}{.231} \\
    \ossshort & \rci{.944}{.918}{.963} & \rci{.951}{.926}{.968} & \rci{.539}{.486}{.591} & \rci{.736}{.691}{.776} \\
    \midrule
    \gptminishort & \rci{.364}{.317}{.414} & \rci{.588}{.538}{.635} & \rci{.581}{.534}{.628} & \rci{.859}{.822}{.889} \\
    \geminifshort & \rci{.991}{.977}{.997} & \rci{.831}{.794}{.863} & \rci{.414}{.369}{.460} & \rci{.756}{.714}{.793} \\
    \qwenshort & \rci{1.00}{.991}{1.00} & \rci{.996}{.984}{.999} & \rci{.633}{.587}{.676} & \rci{.435}{.387}{.485} \\
    \midrule
    \gptshort & \rci{.494}{.443}{.546} & \rci{.580}{.530}{.629} & \rci{.435}{.386}{.486} & \rci{.711}{.666}{.753} \\
    \geminishort & \rci{.976}{.957}{.986} & \rci{.993}{.981}{.998} & \rci{.827}{.789}{.859} & \rci{.978}{.960}{.988} \\
    \glmshort  & \rci{.422}{.377}{.469} & \rci{.910}{.880}{.933} & \rci{.277}{.236}{.323} & \rci{.764}{.721}{.802} \\
    \bottomrule
  \end{tabular}
  \caption{Proportion of reasoning traces that explicitly identify the underlying game as \ttt under each obfuscation. Results are aggregated over $100$ games per model-obfuscation pair, and reported with \wilson.}
  \label{tab:app:obfuscations:ttt_recognition}
\end{table}

The recognition rates in \reftable{tab:app:obfuscations:ttt_recognition} provide other interesting insights.
Recognition is uneven and, unlike optimality, does not increase with model tier: \gpt and \glm identify \hid less often than several smaller models despite matching or exceeding their optimality.
Only \gemini recognizes the game consistently across all obfuscations.

\subsection{\nim Obfuscations}\label{app:obfuscations:nim:rules}

In this section, we describe the game of \nim and introduce its obfuscations.
As with \ttt, each obfuscation preserves the game tree while rewriting the surface representation of states, actions, and winning conditions.

\textbf{\nim} is played on several piles of stones; players alternate turns, and on each turn the active player removes any positive number of stones from a single pile.
We use the mis\`ere variant, in which the player who removes the last stone loses.
Optimal play is governed by the Nim-sum, the bitwise XOR of the pile sizes: a zero Nim-sum is losing, except in endgames where every non-empty pile has size one, which follow a separate parity rule.
Our experiments use the fixed initial configuration $\{1,3,5,7\}$, a forced second-player win.

\textbf{\nimhidlong} (\nimhid) is a pure relabeling.
The piles become four light bulbs and a pile's size becomes a bulb's brightness, so removing stones corresponds to dimming a bulb, and the player who switches off the last light loses.
\textbf{\niminvlong} (\niminv) reverses the surface action while preserving the game tree: instead of removing stones, players add stones to piles until fixed target sizes are reached, and the player who completes the target configuration loses.
\textbf{\nimstorylong} (\nimstory) recasts \nim as a grid escape.
Four characters, one per row, flee a danger advancing from the left; on each turn a player moves one character any number of cells toward the right edge, and the player who helps the last character escape loses, matching the mis\`ere condition.

\subsection{\nim Obfuscation Results}\label{app:obfuscations:nim:results}

Across models and providers, \nim performance is largely stable under the obfuscations.
The \frontier tier remains near-optimal or optimal on every formulation, and most other models stay close to optimal on \nimhid and \nimstory.
\niminv causes the largest degradation, most notably for \oss, \gptmini, and \geminifl.
Even there the drop is modest compared to the more severe failures on some \ttt story formulations in the main text.
This supports the interpretation from \refsupp{app:evals} that \nim is easier for current models than \ttt, since it admits a compact state-value computation based on the Nim-sum, whereas \ttt requires more robust reasoning over spatial layouts and local tactical patterns.

\begin{table}[ht]
  \small
  \centering
  \setlength{\tabcolsep}{2pt}
  \begin{tabular}{lcccc}
    \toprule
    Model & Nim & Hid & Inv & Story \\
    \midrule
    \gptnanoshort & \rci{.955}{.904}{.979} & \rci{.993}{.962}{.999} & \rci{.979}{.940}{.993} & \rci{.987}{.952}{.996} \\
    \geminiflshort & \rci{1.00}{.976}{1.00} & \rci{1.00}{.975}{1.00} & \rci{.954}{.909}{.978} & \rci{1.00}{.975}{1.00} \\
    \ossshort & \rci{.980}{.943}{.993} & \rci{.968}{.926}{.986} & \rci{.917}{.857}{.953} & \rci{.994}{.964}{.999} \\
    \midrule
    \gptminishort & \rci{.993}{.959}{.999} & \rci{1.00}{.976}{1.00} & \rci{.961}{.917}{.982} & \rci{1.00}{.974}{1.00} \\
    \geminifshort & \rci{1.00}{.974}{1.00} & \rci{1.00}{.976}{1.00} & \rci{1.00}{.975}{1.00} & \rci{1.00}{.973}{1.00} \\
    \qwenshort & \rci{1.00}{.974}{1.00} & \rci{1.00}{.975}{1.00} & \rci{.994}{.964}{.999} & \rci{.993}{.960}{.999} \\
    \midrule
    \gptshort & \rci{1.00}{.974}{1.00} & \rci{.993}{.961}{.999} & \rci{1.00}{.973}{1.00} & \rci{.986}{.951}{.996} \\
    \geminishort & \rci{1.00}{.975}{1.00} & \rci{1.00}{.975}{1.00} & \rci{1.00}{.976}{1.00} & \rci{1.00}{.973}{1.00} \\
    \glmshort & \rci{1.00}{.975}{1.00} & \rci{1.00}{.973}{1.00} & \rci{.992}{.957}{.999} & \rci{1.00}{.974}{1.00} \\
    \bottomrule
  \end{tabular}
  \caption{Per-move optimality rates with \wilson for all tiers on \nim and its obfuscations against the \optimal opponent. Each model plays $100$ games per formulation, switching sides after half of the games.}
  \label{tab:app:obfuscations:nim_results}
\end{table}

To inspect whether models explicitly recover the hidden structure, we also compute the proportion of reasoning-trace summaries that mention \nim by name under each obfuscation.

Recognition is high for most models across all three obfuscations, confirming they can usually map the rewritten rules back to \nim.
As with the \ttt results, high recognition does not separate reasoning from recall: once a model identifies the game, it may draw on memorized strategy.
The \niminv column is the informative exception: \gpt and \glm recognize \nim in only $75.7\%$ and $71.3\%$ of traces, yet still play near-optimally.
As on \ttt's \msqadd, competent play persists even when recognition drops, pointing to decision-time reasoning rather than recall.

\begin{table}[H]
  \small
  \centering
  \setlength{\tabcolsep}{2pt}
  \begin{tabular}{lccc}
    \toprule
    Model & Hid & Inv & Story \\
    \midrule
    \gptnanoshort & \rci{.981}{.963}{.990} & \rci{.956}{.933}{.971} & \rci{.956}{.932}{.971} \\
    \geminiflshort & \rci{1.00}{.992}{1.00} & \rci{.936}{.912}{.954} & \rci{.945}{.921}{.962} \\
    \ossshort & \rci{.961}{.939}{.975} & \rci{.956}{.934}{.971} & \rci{.974}{.956}{.985} \\
    \midrule
    \gptminishort & \rci{.907}{.876}{.931} & \rci{.891}{.858}{.917} & \rci{.906}{.875}{.930} \\
    \geminifshort & \rci{1.00}{.993}{1.00} & \rci{.987}{.972}{.993} & \rci{.998}{.989}{1.00} \\
    \qwenshort & \rci{.993}{.979}{.997} & \rci{.988}{.973}{.994} & \rci{.965}{.944}{.978} \\
    \midrule
    \gptshort & \rci{.898}{.861}{.926} & \rci{.757}{.708}{.800} & \rci{.860}{.817}{.895} \\
    \geminishort & \rci{1.00}{.993}{1.00} & \rci{.989}{.976}{.995} & \rci{1.00}{.993}{1.00} \\
    \glmshort  & \rci{.979}{.958}{.989} & \rci{.713}{.664}{.757} & \rci{.937}{.908}{.957} \\
    \bottomrule
  \end{tabular}
  \caption{Proportion of reasoning traces that explicitly identify the underlying game as \nim under each obfuscated formulation. Results are aggregated over $100$ games per model-obfuscation pair and reported with \wilson.}
  \label{tab:app:obfuscations:nim_recognition}
\end{table}

\clearpage
\section{Cost and Reasoning Effort Analysis}\label{app:costs}

This section analyzes the inference cost of the benchmark experiments from \refsection{sec:evals}.
Since all LLM inference runs through remote provider APIs, hardware-level statistics such as GPU type, GPU hours, and energy use are unavailable and would not be comparable across providers.
We therefore report cost through a single observable quantity: output tokens per move, which, given the fixed action format, are dominated by reasoning tokens.

\subsection{Output Tokens Under Obfuscations}\label{app:costs:tokens}

Although per-move optimality is largely stable across obfuscations, output tokens per move increase substantially with surface-form complexity.
This suggests that models often spend additional reasoning effort to translate the obfuscated formulation into a playable state, even when action quality changes only modestly.
\reftable{tab:app:cost:tokens:ttt} and \reftable{tab:app:cost:tokens:nim} report average output tokens per move for all models on \ttt and \nim.

\begin{table}[h]
  \small
  \centering
  \setlength{\tabcolsep}{3pt}
  \begin{tabular}{lccccc}
    \toprule
    Model & TTT & Hid & M.Sq & M.Sq +5 & Story \\
    \midrule
    \gptnanoshort & \mci{7247.8}{578.1} & \mci{8054.9}{592.7} & \mci{10563.6}{779.8} & \mci{12828.4}{743.7} & \mci{15517.9}{954.3} \\
    \geminiflshort & \mci{2245.0}{211.0} & \mci{2485.8}{226.4} & \mci{4529.7}{356.0} & \mci{5766.5}{361.5} & \mci{8825.0}{628.1} \\
    \ossshort & \mci{4220.1}{667.0} & \mci{4335.4}{662.6} & \mci{6147.7}{871.1} & \mci{9446.0}{1128.0} & \mci{6484.3}{698.4} \\
    \midrule
    \gptminishort & \mci{1899.2}{203.4} & \mci{1997.0}{213.9} & \mci{3623.3}{389.2} & \mci{4625.3}{474.3} & \mci{5462.0}{523.3} \\
    \geminifshort & \mci{5068.7}{561.2} & \mci{5282.4}{554.7} & \mci{6825.8}{663.3} & \mci{8879.5}{736.1} & \mci{11961.6}{635.9} \\
    \qwenshort & \mci{2471.2}{197.1} & \mci{2718.5}{242.9} & \mci{2943.8}{228.8} & \mci{4127.6}{267.5} & \mci{4922.7}{203.5} \\
    \midrule
    \gptshort & \mci{806.7}{145.4} & \mci{886.2}{162.0} & \mci{1809.2}{304.2} & \mci{1755.4}{270.7} & \mci{3587.9}{435.8} \\
    \geminishort & \mci{876.2}{100.3} & \mci{955.2}{95.3} & \mci{1484.1}{194.2} & \mci{1819.6}{229.4} & \mci{2171.7}{287.2} \\
    \glmshort & \mci{3771.9}{516.6} & \mci{4193.7}{610.1} & \mci{5472.8}{605.8} & \mci{12327.7}{1325.5} & \mci{13867.7}{1100.8} \\
    \bottomrule
  \end{tabular}
  \caption{Average output tokens per move for all tiers on \ttt and its obfuscations. Values are reported as means with \student.}
  \label{tab:app:cost:tokens:ttt}
\end{table}

The token trends differ from the optimality trends.
For \ttt, output tokens generally increase with obfuscation complexity, from the standard game through \hid, \msq, \msqadd, and \story.
For \nim the ordering is less strict, but \nimhid and \nimstory still require more tokens than the standard game, while \niminv is typically the cheapest, at least for the \intermediate and \efficient tier models.

\begin{table}[ht]
  \small
  \centering
  \setlength{\tabcolsep}{3pt}
  \begin{tabular}{lcccc}
    \toprule
    Model & Nim & Hid & Inv & Story \\
    \midrule
    \gptnanoshort & \mci{11662.5}{885.8} & \mci{13745.3}{863.2} & \mci{10456.7}{700.5} & \mci{11875.5}{885.8} \\
    \geminiflshort & \mci{4804.2}{274.9} & \mci{7588.7}{302.7} & \mci{5652.2}{304.0} & \mci{8236.8}{362.7} \\
    \ossshort & \mci{4862.7}{358.3} & \mci{6074.5}{401.6} & \mci{4122.5}{290.6} & \mci{5482.3}{344.8} \\
    \midrule
    \gptminishort & \mci{3069.8}{221.5} & \mci{3893.5}{272.3} & \mci{2942.6}{210.1} & \mci{3927.1}{291.2} \\
    \geminifshort & \mci{6693.5}{365.5} & \mci{8557.7}{405.0} & \mci{6059.9}{328.6} & \mci{7987.7}{369.4} \\
    \qwenshort & \mci{3213.4}{226.4} & \mci{4621.4}{186.6} & \mci{3893.3}{198.7} & \mci{4732.8}{219.9} \\
    \midrule
    \gptshort & \mci{2025.8}{239.1} & \mci{1865.2}{229.1} & \mci{2286.9}{279.4} & \mci{2155.3}{255.7} \\
    \geminishort & \mci{706.5}{40.5} & \mci{848.0}{50.0} & \mci{1344.1}{74.3} & \mci{1116.1}{83.6} \\
    \glmshort & \mci{5234.1}{597.1} & \mci{5495.4}{644.6} & \mci{6044.8}{687.7} & \mci{6665.7}{695.3} \\
    \bottomrule
  \end{tabular}
  \caption{Average output tokens per move for all tiers on \nim and its obfuscations. Values are reported as means with \student.}
  \label{tab:app:cost:tokens:nim}
\end{table}

Absolute token counts vary substantially and do not track model tier: the \efficient-tier \gptnano is the heaviest of all, while the \frontier tier spans the full range, with \gemini and \gpt the most efficient and \glm among the heaviest.
Similar move-level optimality can therefore correspond to very different inference costs.
\gemini is the clearest case, playing \nim optimally (\refsupp{app:evals}) with about a third of \gpt's tokens per move, suggesting a more direct computation of the Nim-sum, while on \ttt the two are comparable.
Therefore the gap is likely task-specific.

\subsection{Effect of Reasoning Effort}\label{app:costs:reasoning}

We additionally examine how reasoning effort affects play and token cost for \gptnano and \gptmini on \ttt against the \optimal opponent; the setup otherwise matches \refsection{sec:experiments:setting}.
\reftable{tab:app:cost:optimality:reasoning:ttt} reports per-move optimality and output tokens per move at low, medium, and high effort.

\begin{table}[h]
    \small
    \centering
    \setlength{\tabcolsep}{5pt}
    \begin{tabular}{lcccc}
        \toprule
        \multirow{2}[3]{*}{Model} & \multirow{2}[3]{*}{Metric} & \multicolumn{3}{c}{Reasoning Effort} \\
        \cmidrule(lr){3-5}
        & & Low & Medium & High \\
        \midrule
        \multirow{2}[3]{*}{\gptnanoshort} & Optimality & \rci{.731}{.667}{.786} & \rci{.843}{.785}{.887} & \rci{.880}{.828}{.918} \\
        & \makecell[c]{Tokens \\ per move} & \mci{1055.4}{78.7} & \mci{3526.6}{293.0} & \mci{7247.8}{578.1} \\
        \midrule
        \multirow{2}[3]{*}{\gptminishort} & Optimality & \rci{.758}{.692}{.813} & \rci{.887}{.836}{.923} & \rci{.922}{.878}{.952} \\
        & \makecell[c]{Tokens \\ per move} & \mci{351.1}{32.5} & \mci{762.9}{73.2} & \mci{1899.2}{203.4} \\
        \bottomrule
    \end{tabular}
    \caption{Per-move optimality and output tokens per move on standard \ttt against the \optimal opponent for the two GPT models under low, medium, and high reasoning effort. Optimality is reported with \wilson and token means with \student.}
    \label{tab:app:cost:optimality:reasoning:ttt}
\end{table}

Confirming our intuition, higher reasoning effort improves optimality for both GPT models while also raising their output-token count.
Moving from low to high effort increases optimality by roughly $15$ percentage points, but the output tokens grow by a factor of $5$ for \gptmini and $7$ for \gptnano.
Higher effort therefore pushes play closer to optimal, but at a large inference-cost multiplier.

\clearpage
\section{\recap Details}\label{app:recap}

This supplementary section gives the architectural details of \recap omitted from \refsection{sec:recap} for brevity.
We recall only what is needed: \recap keeps the planner-executor loop and case memory of \baseline and adds the self-reflector and rule extractor, together with an auxiliary state descriptor.

\subsection{Decision Loop}\label{app:recap:decision_loop}

The planner and executor communicate through structured JSON schemas.
At each move the planner emits either a plan or a final answer.
A plan is a list of subtasks; the executor solves them one at a time, each time receiving the current assignment together with the results of the subtasks already solved for that plan.
Once all subtasks are answered, control returns to the planner, which again emits a plan or a final answer.
The loop runs for at most $3$ iterations; if no final answer is produced within this budget, or a message fails to parse, the move is recorded as a failure and the game terminates.

Alongside the final answer, the planner emits a natural-language action intent describing the intended effect of the chosen action.
The self-reflector later uses this intent to judge whether the action achieved what the planner intended.

\subsection{Reflection and Rule Memory}\label{app:recap:reflection_and_rule_memory}

Once an episode ends, the self-reflector reviews the whole trajectory in a single pass and produces one reflection per move, replacing the uniform terminal reward with a finer, per-move signal (\refsection{sec:recap}).
Drawing on the state, the played action and its stated intent, and the opponent's reply, each reflection judges the move along two axes: its local quality, whether it was sound at the point it was played, and its contribution to the final outcome.
It also assesses whether the action achieved its stated intent, and, where relevant, notes the turning point at which the game was lost and records a short lesson for future play.

This per-move judgement also determines how a retrieved case is shown to the planner: a locally sound move that did not harm the outcome is presented as an example to emulate, and any other move as a mistake to avoid.
Because the label is assigned per move, a sound move from a lost game is still shown as a positive example rather than penalized for the eventual loss.
Which cases are retrieved in the first place is balanced separately (\refsupp{app:recap:retrieval}).

The rule extractor is what lets \recap generalize beyond the states it has actually encountered (\refsection{sec:recap}).
Beyond the periodic, self-correcting regeneration described there, two constraints shape the rule set.
It is kept deliberately small, on the order of twenty rules, and biased toward rules supported by several cases or by clear win/loss patterns, so that one-off observations do not accumulate into noise.
More importantly, the extractor is required to phrase rules as general strategic principles rather than statements about particular board positions; this position-independence is precisely what allows them to apply to unseen states, complementing case retrieval, which can only surface states resembling those already stored.
The set is presented as a short list in the planner's prompt.

\subsection{Retrieval}\label{app:recap:retrieval}

To reuse past experience, the agent needs a key by which to compare the current position against stored cases.
The natural choice is the state observation itself: it is exactly what the environment exposes during play and needs no extra model call.
At each move this key is embedded with a sentence encoder\footnote{\texttt{princeton-nlp/sup-simcse-bert-base-uncased}}~\cite{simcse} truncated to $256$ tokens, and the $k=8$ most similar cases are retrieved by cosine similarity.

\subsubsection{Stratified Retrieval}\label{app:recap:retrieval:stratified}

Ranking by similarity alone can return a neighborhood dominated by a single outcome, for example, only cases from lost games, starving the planner of contrasting examples.
Stratified retrieval counters this by splitting the stored cases into a positive and a non-positive stratum by their recorded reward and drawing roughly half of the $k$ slots from each.
If one stratum cannot fill its share, the remaining slots are backfilled with the next most similar cases regardless of stratum.
The retrieved set therefore holds both successful and cautionary cases whenever both exist near the query.

\subsection{State Descriptor}\label{app:recap:state_descriptor}

Using the raw observation as the key, i.e. the state mode, makes retrieval sensitive to surface form.
Two positions that are strategically identical but written differently, as under the obfuscations of \refsection{sec:evals}, embed to different keys and fail to match.
Similarly, if the game engine does not impose a canonical ordering of the board positions (or the state representation in general), two different observations also map to different embeddings.
This motivates describing the state in natural language before embedding, decoupling retrieval from the raw encoding.
That is the role of the state descriptor, which is introduced in this section.

\begin{figure}[ht]
    \centering
    \includegraphics[width=0.9\linewidth]{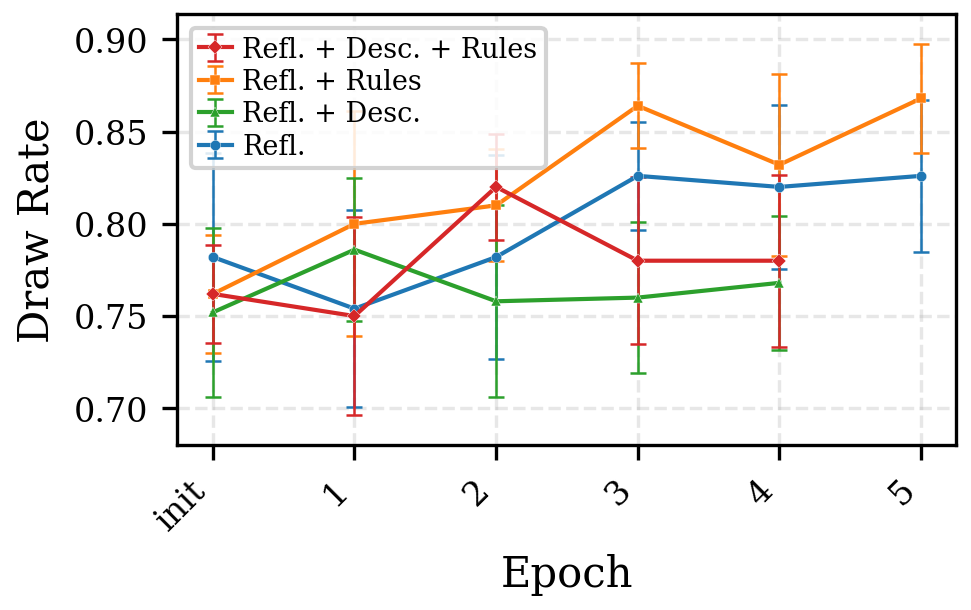}
    \caption{Effect of the state descriptor on \ttt draw rate with \gptnano at medium reasoning effort against the \optimal opponent. State-mode variants (\textit{Refl.}, \textit{Refl. + Rules}) use the raw observation as the retrieval key; the corresponding description-mode variants (\textit{Refl. + Desc.}, \textit{Refl. + Desc. + Rules}) replace it with the generated state description. Lines show means over $10$ independent runs; error bars show \student.}
    \label{fig:app:recap:ablation:draw_rate_descriptions}
\end{figure}

Concretely, the state descriptor is a dedicated model call that reads the raw observation and returns a short natural-language description of the current position, which is then embedded in place of the raw observation, the description-mode retrieval key.
The aim is to make the key independent of surface form or canonical ordering, so that positions that are strategically equivalent but encoded differently map to nearby keys; the price is an extra model call on every move.

\subsubsection{Ablation Study}\label{app:recap:state_descriptor:ablation}

We evaluate description mode in the same setting as \refsection{sec:experiments}, \gptnano at medium effort against the \optimal opponent on \ttt, layering it on both the reflection-only agent and the reflection-with-rules agent, and comparing against the corresponding variants with state embeddings.
\reffigure{fig:app:recap:ablation:draw_rate_descriptions} reports the draw rate across the five learning epochs.

Description mode does not help.
Both description-mode variants plateau below their state-mode counterparts.
To see why, we measured the retrieval success rate, i.e. how often a case whose state was already in memory was actually retrieved for that state.
Under state mode, this is effectively $100\%$, since an embedded raw observation matches its own stored key exactly.
Under description mode it falls below $100\%$, because the natural-language descriptions of the same states are no longer identical or sufficiently close in the embedding space, so the matching case is no longer reliably retrieved among the top $k$.

\subsection{Component Setup}\label{app:recap:component_setup}

\recap's components are exposed as independent switches, which is what makes the ablations of \refsection{sec:experiments:ablation} possible: reflection, rule extraction, and the state descriptor can each be turned on or off without changing the rest of the pipeline.
Beyond this, the framework allows a different LLM backend per component and exposes the memory and retrieval hyperparameters discussed above.
In all main-text experiments we use a single model for every component, matched to the player under evaluation, so that any improvement is attributable to the method rather than to a stronger auxiliary model; the framework nonetheless supports arbitrary model combinations.

\clearpage
\section{Relation to OpenSpiel 2.0}\label{app:openspiel}

Concurrently with this work, OpenSpiel~2.0~\cite{openspiel2} introduced structured state, observation, and action representations that serialize losslessly to JSON, together with a minimax solver for small two-player zero-sum perfect-information games.
These additions are motivated explicitly by ease of use with large language models, and overlap in spirit with \omniplay's interface adapters; we clarify here how the two relate.
\omniplay predates and is independent of these features, and remains necessary for the study conducted in this paper.
It still uses OpenSpiel's engine and game implementations, but complements it with features that are tailored for LLM play and studying LLMs and differ in four respects.

\paragraph{Representation}
OpenSpiel~2.0 serializes its structs to a JSON format and sends them, in that form, to a frontend, a database, or a language model.
LLM Adapter in \omniplay instead renders the state and legal actions as natural-language text tuned for prompting, layered on top of the same OpenSpiel engine.
These are different design points rather than one dominating the other.
A JSON encoding is unambiguous, lossless, compact, and trivially machine-parsed, which simplifies tool-based agent integration, logging, and reproducibility.
A natural-language encoding mirrors how a game's rules would be explained to a person and provides the surface on which the obfuscations below can act, making it the more natural substrate for our goal of probing whether a model understands and reasons about rules rather than recognizes a serialized format.

\paragraph{Obfuscations}
While OpenSpiel's structs provider faithful state representation, \omniplay's adapter is designed to support game rule obfuscations: semantics-preserving surface transforms (\refsection{sec:evals:obfuscations}) that alter the natural-language representation while leaving the game tree unchanged.
This mechanism presents the same game under rules the model is unlikely to have memorized, and is the basis for disentangling memorization from reasoning.

\paragraph{Stochastic optimal opponent}
OpenSpiel~2.0's minimax solver is an oracle: it computes the minimax value of every action in each state and exposes these action values, from which the set of optimal moves can be recovered, but it neither defines a playing policy nor prescribes how ties among optimal moves are resolved.
\omniplay instead wraps \minimax into a ready-to-use \optimal opponent that, at each state, samples uniformly among all moves attaining the optimal value.
This stochastic tie-breaking is what our per-move optimality metric consumes: the metric excludes states where every legal move is optimal and otherwise credits any move \minimax could have selected (\refsection{sec:evals}), and the sampling additionally exposes the player to a diverse distribution of optimal continuations rather than a single fixed line.

\paragraph{Evaluation harness}
Finally, OpenSpiel provides evaluation primitives rather than an end-to-end study harness.
Its \texttt{evaluate\_bots} routine plays a single game between a list of bots and returns the terminal utilities, and OpenSpiel~2.0 adds population-level rating aggregation (Elo and Soft-Condorcet) together with an example server through which one language model can play interactively.
It ships no driver for running many games across players and rule formulations, no parallelization of such matchups, and no per-move analysis.
\omniplay supplies exactly this layer: through its Interface Adapter it exposes a single unified interface over heterogeneous players, from random and algorithmic baselines to search-based opponents and LLMs, then orchestrates repeated games with side switching, parallelizes them across players and rule formulations, and scores play at the per-move level against the stochastic \optimal opponent, aggregating with the \wilson reported in our tables.
Our metric therefore evaluates whether each individual move was optimal, a finer-grained signal than the terminal outcomes and population ratings that OpenSpiel's evaluation utilities compute.
This full pipeline, rather than any single adapter, is what the study in this paper relies on.

None of these distinctions are exclusive.
Because the obfuscation mechanism is largely independent of the encoding, it could equally be applied to OpenSpiel's structs, for instance by relabeling actions and remapping observation fields under a semantics-preserving transform.
The two frameworks are therefore complementary and could be merged: \omniplay's obfuscations, stochastic \optimal opponent, and evaluation orchestration could be built directly on OpenSpiel~2.0's structured representations, pairing the machine-readability of JSON with our reasoning-versus-recall probes.
We accordingly view OpenSpiel~2.0 as complementary infrastructure rather than a substitute for \omniplay, and see extending or merging the two as a natural direction for future work.

\end{document}